# Deciding When Not to Decide: Indeterminacy-Aware Intrusion Detection with NeutroSENSE

Eyhab Al-Masri
School of Engineering and Technology
University of Washington
Tacoma, WA, USA
ealmasri@uw.edu

*Abstract*— **This paper presents NeutroSENSE, a neutrosophic-enhanced ensemble framework for interpretable intrusion detection in IoT environments. By integrating Random Forest, XGBoost, and Logistic Regression with neutrosophic logic, the system decomposes prediction confidence into truth (T), falsity (F), and indeterminacy (I) components, enabling uncertainty quantification and abstention. Predictions with high indeterminacy are flagged for review using both global and adaptive, class-specific thresholds. Evaluated on the IoT-CAD dataset, NeutroSENSE achieved 97% accuracy, while demonstrating that misclassified samples exhibit significantly higher indeterminacy (I = 0.62) than correct ones (I = 0.24). The use of indeterminacy as a proxy for uncertainty enables informed abstention and targeted review—particularly valuable in edge deployments. Figures and tables validate the correlation between I-scores and error likelihood, supporting more trustworthy, human-in-the-loop AI decisions. This work shows that neutrosophic logic enhances both accuracy and explainability, providing a practical foundation for trust-aware AI in edge and fog-based IoT security systems.**

## I. Introduction

Intrusion Detection Systems (IDS) are pivotal in protecting digital infrastructures against the growing sophistication of cyberattacks, particularly within cloud and IoT-enabled environments [1]. However, the dynamic and unpredictable nature of contemporary threats — including zero-day vulnerabilities and polymorphic malware — often renders traditional IDS approaches inadequate [2]. Conventional machine learning-based IDS typically operate under the assumption of clean, deterministic data and binary classification boundaries, limiting their effectiveness in real-world scenarios characterized by uncertainty, inconsistency, and ambiguous decision boundaries [3].

While anomaly-based detection systems are effective at spotting previously unseen threats, they often experience high false positive rates because they lack mechanisms to account for uncertainty in their predictions [4]. For instance, an IDS that assigns 51% probability to the "attack" class and 49% to "normal" may still trigger an alert with full confidence — even though the evidence is inconclusive and likely unsuitable for automated response. This necessitates a paradigm shift toward more interpretable and trust-aware AI systems that can incorporate the degrees of uncertainty inherent in network traffic data and model decision ambiguity more effectively [5].

To improve generalization and reduce overfitting to specific attack patterns, many recent IDS approaches have adopted ensemble learning techniques, which combine multiple base models to provide more stable and robust predictions across diverse intrusion types [6]. These methods, which combine multiple base learners — such as decision trees, gradient boosting models, or logistic regression — have consistently demonstrated superior performance in complex and high-dimensional cybersecurity datasets. By aggregating diverse classifiers, ensemble-based IDS can capture a broader spectrum of attack behaviors and offer improved accuracy over single-model approaches. However, despite these advantages, ensemble methods often operate as black boxes, producing consensus outputs without transparent reasoning or interpretability [7]. In high-stakes domains like cybersecurity, where decisions may trigger automated defenses or human escalation, this lack of explainability becomes a critical bottleneck. Moreover, ensemble models typically collapse prediction disagreement into a single output score, masking cases of model conflict or low consensus, which could otherwise serve as valuable indicators of uncertainty or anomalous behavior [8].

A critical shortcoming in most existing intrusion detection models lies in their inability to quantify and reason about the indeterminacy present in prediction outcomes [4]. Traditional supervised classifiers tend to produce hard labels or scalar confidence scores, which fail to capture the ambiguity in cases where the input exhibits characteristics of both benign and malicious traffic. This ambiguity is particularly prevalent in noisy, incomplete, or overlapping feature spaces — a common reality in IoT and cloud-based environments. Even ensemble learning techniques, which improve overall accuracy by aggregating diverse models, often collapse uncertainty into average votes or probabilistic scores without explicitly modeling the degree of uncertainty or flagging samples that may require further inspection [9]. As a result, these systems are unable to effectively identify and respond to IoT or edge computing cases that may signal stealthy or emerging threats.

To address the challenges of ambiguity and uncertainty in intrusion detection, neutrosophic logic offers a compelling foundation. Introduced by Florentin Smarandache [10], neutrosophic sets extend classical logic by introducing three independent components for each decision: truth (T), falsity (F), and indeterminacy (I). Unlike fuzzy or probabilistic models that reduce uncertainty to a single scalar, neutrosophic reasoning explicitly models indeterminacy, preserving ambiguity [10]. This makes it well-suited for domains like intrusion detection, where borderline behavior is common and critical to detect. Neutrosophic logic has also been applied in uncertain classification to enhance interpretability, offering a strong foundation for trust-aware AI [11].

In this paper, we introduce NeutroSENSE, a neutrosophic-enhanced ensemble framework for interpretable and uncertainty-aware intrusion detection in IoT environments. The main contributions of this work are:

- **Neutrosophic Integration in Ensemble Classification:** We augment an ensemble of Random Forest, XGBoost, and Logistic Regression with neutrosophic logic, computing T, I, and F scores per prediction to enhance interpretability and quantify uncertainty.
- **Indeterminacy-Aware Flagging Mechanism**: We flag high-indeterminacy samples—those with near-equal support for multiple classes—for abstention or human review, forming the basis of a lightweight, trust-aware decision mechanism.
- **Adaptive Thresholding Strategy**: We design class-specific thresholds that adjust dynamically based on indeterminacy, allowing NeutroSENSE to abstain in uncertain cases while preserving high coverage.
- **Empirical Validation on IoT Intrusion Data**: We evaluate NeutroSENSE on filtered IoT intrusion data, achieving 97% accuracy and flagging over 12,000 high-indeterminacy cases—many aligned with misclassifications—demonstrating the effectiveness of our uncertainty-aware framework.
- **Toward Trust-Aware AI in Cybersecurity**: We contribute to explainable, trust-aware AI by showing how uncertainty quantification improves decision transparency and safety in edge-based cybersecurity environments.

## II. RELATED WORK

Intrusion Detection Systems (IDS) have long been a focal point in cybersecurity research, especially with the abundance of Internet of Things (IoT) devices and the inherent uncertainty in real-time traffic analysis [6]. This section reviews three major strands of literature relevant to our work: (1) ensemble learning in IDS, (2) handling uncertainty and indeterminacy using neutrosophic logic, and (3) integration of explainable and abstaining classifiers.

### A. Ensemble Learning for Intrusion Detection

Ensemble-based classifiers have been widely adopted in IDS to improve accuracy and robustness against noisy or imbalanced data [6]. Bagging and boosting techniques such as Random Forest, AdaBoost, and hybrid stacking models have demonstrated significant improvements over standalone classifiers [12]. More recent approaches have explored dynamically weighted ensembles and cost-sensitive voting schemes [13]. However, most ensemble models produce deterministic outputs and fail to expose model uncertainty, which limits their usefulness in high-stakes environments like IoT security [14].

### B. Uncertainty Modeling with Neutrosophic Logic

To address the limitations of conventional fuzzy and probabilistic models, researchers have proposed Neutrosophic Logic (NL), a generalization of classical and fuzzy logic that introduces an explicit indeterminacy component alongside truth and falsity [12]. Neutrosophic classifiers have been shown to outperform fuzzy and intuitionistic fuzzy systems, especially in noisy, imbalanced, or adversarial data conditions. For instance, Akbulut et al. proposed NWELM—a Neutrosophic Weighted Extreme Learning Machine—which uses neutrosophic c-means to weight instances by their T, I, F scores, significantly improving classification on benchmark datasets [15].

Another key contribution is the use of interval neutrosophic sets in ensemble neural networks, where paired networks estimate both truth and falsity degrees, and indeterminacy is derived as their complement [12]. These methods not only boost detection accuracy but also provide explainable abstention in uncertain regions—an important feature for human-in-the-loop systems.

### C. Uncertainty Modeling with Neutrosophic Logic

Several systems have incorporated NL directly into IDS design. Kavitha et al. introduced a Neutrosophic Logic Classifier that partitions traffic into three parts—normal, abnormal, and indeterministic—based on thresholds in T, I, and F scores [13]. Their work showed promising reductions in false positives and better handling of borderline traffic [13]. However, their models were optimized using genetic algorithms and lacked modern ensemble learners or abstention-aware evaluation.

In contrast, our work uniquely integrates neutrosophic indeterminacy estimation within an ensemble classification framework (Random Forest, XGBoost, Logistic Regression), and introduces a class-conditional adaptive thresholding mechanism to selectively abstain on highly indeterminate predictions. This supports human review in high-risk predictions and aligns with explainable AI (XAI) principles.

### D. Comparison with Prior Work

Table 1 summarizes key differences between our approach and relevant existing intrusion detection systems leveraging ensemble learning and neutrosophic logic. Unlike earlier methods that apply rule-based reasoning, shallow networks, or fixed classifiers [12, 13, 18], our proposed NeutroSENSE framework uniquely combines modern ensemble learners—Random Forest, XGBoost, and Logistic Regression—with neutrosophic logic for uncertainty modeling. To our knowledge, this is the first framework to integrate adaptive abstention thresholds driven by indeterminacy scores, enabling interpretable and trust-aware decision-making in ambiguous or high-risk intrusion scenarios. Prior works often lack either uncertainty quantification, interpretability, or scalability to IoT environments [6]. Our method addresses all three by quantifying prediction ambiguity (T, I, F), supporting abstention, and validating performance on a realistic IoT dataset (IoT-CAD) [16], distinguishing it from black-box or taxonomy-focused approaches.

TABLE I. PRIOR WORK COMPARISON

| Method | Ensemble Learning | Uncertainty Modeling | Abstention / Flagging | explainable AI (XAI) | Dataset Scope | Key Limitation |
|---|---|---|---|---|---|---|
| Kraipeerapun et al. (2007) [12] | Bagging + NN pairs (T/F) | ≈ | ✓ | ≈ | UCI datasets | Shallow NN, limited scale |
| Kavitha et al. (2012) [13] | Rule-based with GA optimization | ✓ | ✓ | ✓ | KDD Cup 99 | No statistical ensembles |
| Akbulut et al. (2019) [15] | NWELM + NCM weighting | ✓ | × | × | Synthetic, KEEL | No abstention or interpretability |
| Elhassouny et al. (2019) [17] | NS hybrid taxonomy (review) | ✓ | ≈ | ≈ | N/A | Lacks IoT and experimental focus |
| Chebrolu et al. (2005) [18] | BN + CART hybrid ensemble | × | × | × | KDD Cup 99 | No uncertainty modeling |
| Wang et al. (2010) [19] | Fuzzy clustering + ANN | ≈ | × | × | KDD Cup 99 | Lacks explicit indeterminacy |
| Idhammad et al. (2023) [20] | NB + Random Forest | × | × | × | CIDDS-001 | Black-box, no uncertainty |
| Nguyen et al. (2023) [21] | Fog-Cloud hierarchical model | × | × | ≈ | CAIDA, KDD Cup 99, NB15 | No trust modeling or abstention |
| Kraipeerapun et al. (2009) [22] | T/F NN bagging w/ interval NS | ✓ | ✓ | ✓ | UCI ML datasets | Not IDS-specific datasets |
| Liao et al. (2013) [23] | Review of IDS taxonomies | × | × | × | N/A | No empirical methods |
| **Our NeutroSENSE** | RF + XGBoost + LR (neutrosophy) | ✓ | ✓ | ✓ | IoT-CAD | Ensemble w/ neutrosophy & abstention |

## III. SYSTEM ARCHITECTURE AND METHODOLOGY

### A. Overview of the NeutroSENSE Framework

The proposed NeutroSENSE (Neutrosophic Sensing for Security and Explainability) framework is a novel intrusion detection architecture designed for real-time, trustworthy classification in edge and fog computing environments. It integrates ensemble learning with neutrosophic logic to explicitly quantify prediction indeterminacy, allowing the system to abstain from decisions in ambiguous cases and support human-in-the-loop validation. As shown in Figure 1, NeutroSENSE is modular and lightweight, supporting scalable deployment across distributed edge gateways while enabling high interpretability, making it ideal for low-latency and mission-critical security scenarios in IoT ecosystems.

The framework consists of four core layers: (1) a Data Collection module that collects data from IoT and network edge nodes; (2) a Preprocessing module for normalization and feature encoding; (3) a Neutrosophic Ensemble Classifier comprising Random Forest, XGBoost, and Logistic Regression models; and (4) a Neutrosophic Reasoning module that computes Truth (T), Falsity (F), and Indeterminacy (I) scores. Based on adaptive thresholds, the system either outputs a classification (Normal or Malicious) or flags the input as Indeterminate, forwarding it to the Review Queue. A Logging module captures all outcomes for audit and retraining, forming a feedback loop to improve model robustness. This modular architecture enables real-time, interpretable intrusion detection on edge gateways in distributed IoT and fog environments.

This architectural design directly addresses the limitations observed in prior work. Traditional IDS approaches rely on deterministic predictions with no mechanism to flag uncertainty [10]. While ensemble models have improved robustness, they often obscure disagreement between base learners. Neutrosophic logic, which decomposes predictions into truth (T), falsity (F), and indeterminacy (I) components, offers a principled solution. However, previous neutrosophic-based IDS systems either employed single learners or lacked integration with modern ensemble methods and adaptive abstention strategies. NeutroSENSE bridges this gap, unifying all three elements—neutrosophic reasoning, ensemble classification, and adaptive abstention—into one interpretable and edge-deployable system.

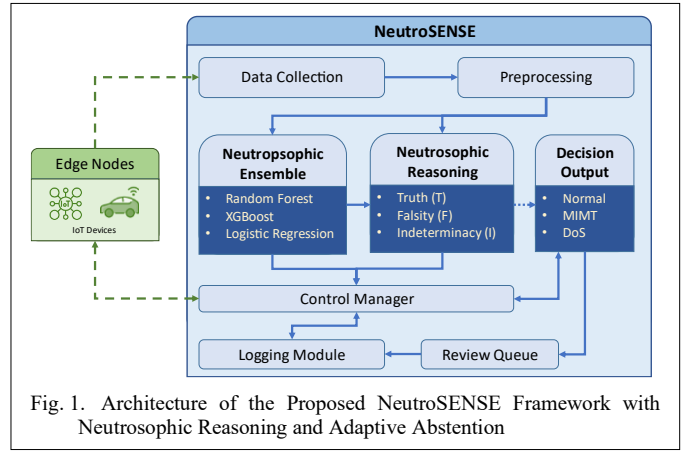

Fig. 1. Architecture of the Proposed NeutroSENSE Framework with Neutrosophic Reasoning and Adaptive Abstention

### B. Architectural Layers and Components

1. **Edge Nodes and Data Collection**: Edge nodes, including IoT devices and gateways, form the first layer, capturing traffic, system logs, and metadata. Due to limited edge resources, the system prioritizes early feature extraction and lightweight data representation for efficient on-device processing.
2. **Preprocessing Module**: Incoming data is cleaned and standardized using one-hot encoding, normalization, and optional feature selection. This ensures consistent input while maintaining low-latency responsiveness critical to real-time environments.
3. **Neutrosophic Ensemble Classifier**: An ensemble of Random Forest, XGBoost, and Logistic Regression outputs class probabilities, chosen for their speed, interpretability, and suitability for imbalanced data. These are combined and passed to the neutrosophic reasoning layer.
4. **Neutrosophic Reasoning Module**: This module decomposes predictions into T (truth), F (falsity), and I (indeterminacy). An adaptive thresholding mechanism flags high-indeterminacy (I) predictions for abstention, enabling uncertainty-aware decisions beyond standard classification.

5. **Control Manager:** This module orchestrates real-time prediction by aggregating class probabilities and forwarding them for neutrosophic decomposition, enabling swift, on-device decision-making.
6. **Decision Output and Review Queue**: Predictions exceeding class-specific indeterminacy (*I*) thresholds are flagged and sent to a review queue; others are labeled as normal or malicious. This selective abstention supports safe automation.
7. **Logging Module and Feedback Integration**: All outcomes—including predictions, abstentions, and manual reviews—are logged to support retraining, threshold refinement, and long-term system improvement through expert feedback.

*C. Design Rationale and Deployment Context*

NeutroSENSE addresses the need for interpretable, context-aware AI in IoT and edge cybersecurity. By enabling indeterminacy-aware abstention and flagging low-confidence cases, it supports Responsible AI. Its lightweight design allows fast local decisions, escalating only uncertain cases to the cloud or human review—reducing bandwidth, latency, and central dependency. In this context, NeutroSENSE is well-suited for integration with edge or fog gateways positioned closer to the points of data generation.

Furthermore, by coupling neutrosophic logic with abstention-aware decisions, NeutroSENSE introduces a novel paradigm for trustworthy AI in edge security. Its explicit modeling of indeterminacy and transparent abstention supports deployment in smart cities, critical infrastructure, and autonomous IoT. The design aligns with explainable AI, federated learning, and resilient edge computing—addressing a key gap in cybersecurity: knowing when not to decide.

## IV. EVALUATION AND RESULTS

We evaluated the proposed NeutroSENSE framework using the IoT-CAD dataset developed at UNSW Canberra [16]. Specifically, we used the Linux variant of the processed dataset, titled *lfiltered_data_Attribution.csv*, which includes rich behavioral and system-level telemetry along with multi-label annotations for what (attack type), how (execution method), and why (attacker intent). For this study, we focused on the what label, representing a multi-class intrusion detection task. After filtering out rare classes with insufficient samples, the final dataset included over 140,000 labeled samples spanning diverse attack categories such as DDoS, MITM, Injection, and Probing. We applied a multi-step preprocessing pipeline to prepare the dataset. Label encoding converted intrusion categories into numeric labels. A 20% holdout set was reserved to assess generalization, preserving class distribution. The remaining 80% was balanced via the Synthetic Minority Over-sampling Technique (SMOTE) to deal with class imbalance [24].

Furthermore, features were then normalized using StandardScaler, and missing values were handled through zero imputation. For model training, we used three classifiers: a Random Forest with *log2* feature selection, depth constraints, and class-weight balancing; XGBoost with regularization terms ($\gamma$, $\alpha$, $\lambda$), subsample tuning, and column sampling; and a Logistic Regression model with L2 regularization and balanced class weights.

All models were evaluated on two distinct test sets: (1) a held-out validation set representing the original data distribution prior to SMOTE balancing, used to assess generalization to real-world class proportions, and (2) a post-SMOTE stratified test set, drawn from the oversampled training data, used for deeper analysis of ensemble performance and neutrosophic scoring.

To ensure statistical robustness and minimize class imbalance, we excluded rare categories with fewer than two samples prior to training. Table II presents the number of samples retained per class after filtering, providing a clear view of the dataset composition used for both SMOTE-based training and generalization assessment.

TABLE II. CLASS DISTRIBUTION OF THE FILTERED DATASET USED FOR NEUTROSENSE EVALUATION.

| Class | Sample Count |
|---|---|
| Normal | 75478 |
| Man-in-the-Middle (MITM) | 57825 |
| Injection | 27946 |
| Denial of Service (DoS) | 24124 |
| Distributed Denial of Service (DDoS) | 23484 |
| Password | 20711 |
| Malware | 20566 |
| Probing | 17212 |

Table II class distribution in the filtered dataset was used for training and evaluation. Rare classes were removed prior to resampling, ensuring a balanced and representative multi-class intrusion detection task.

*A. Base Model Performance*

To establish a performance baseline, we evaluated each individual classifier—Random Forest (RF), XGBoost (XGB), and Logistic Regression (LR)—on the holdout test set, which represents the original (pre-SMOTE) class distribution. This allows us to assess the generalization capability of each model in a realistic scenario, as demonstrated in Table III.

TABLE III. CLASSIFICATION ACCURACY.

| Model | Accuracy (Holdout) |
|---|---|
| Random Forest (RF) | 97.08% |
| XGBoost (XGB) | 97.97% |
| Logistic Regression (LR) | 80.96% |

A more detailed evaluation of XGBoost is provided in Table IV, showing class-wise precision, recall, and F1-score. The model achieved near-perfect scores for most classes, particularly MITM, malware, and password attacks, all exceeding 0.99 F1-score. The only notable drop occurred in detecting Normal traffic, which showed slightly reduced recall due to misclassification as other benign-seeming categories.

The confusion matrix in Figure 2 highlights that most errors stem from normal traffic being misclassified as benign attack types, which is a known challenge in anomaly-based IDS systems. While all base learners were evaluated, we present

XGBoost's confusion matrix as it was the top-performing model, and the others exhibited similar patterns.

TABLE IV. XGBOOST CLASSIFICATION PERFORMANCE ON HOLDOUT SET.

| Class | Precision | Recall | F1-Score |
|---|---|---|---|
| Normal | 0.99 | 0.94 | 0.96 |
| DDoS | 0.97 | 0.99 | 0.98 |
| DoS | 0.99 | 0.99 | 0.99 |
| Injection | 0.94 | 1.00 | 0.97 |
| Malware | 0.98 | 1.00 | 0.99 |
| MITM | 0.99 | 1.00 | 0.99 |
| Password | 0.98 | 1.00 | 0.99 |
| Probing | 0.98 | 1.00 | 0.99 |
| **Macro Avg.** | **0.98** | **0.99** | **0.98** |

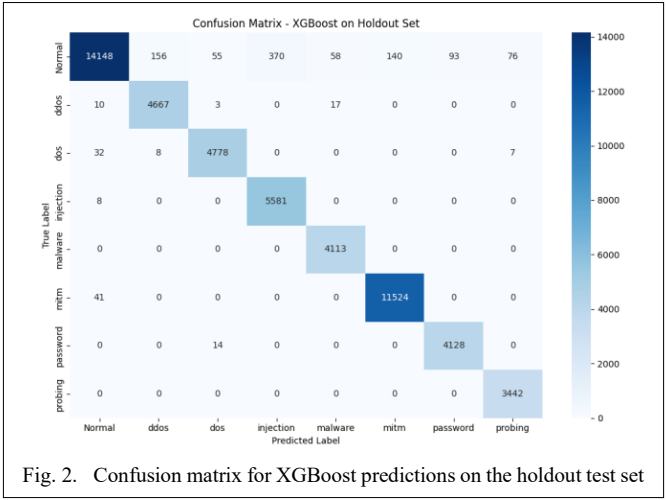

Fig. 2. Confusion matrix for XGBoost predictions on the holdout test set

### B. Ensemble Classification Results (Pre-Abstention)

To assess the performance of the ensemble without indeterminacy handling, we combined predictions from Random Forest, XGBoost, and Logistic Regression by averaging their class probability distributions and selecting the class with the highest combined confidence. This soft voting strategy provides a unified view of the model consensus without incorporating abstention mechanisms.

TABLE V. PERFORMANCE OF THE ENSEMBLE CLASSIFIER USING SOFT VOTING (PRE-NEUTROSOPHIC SCORING).

| Class | Precision | Recall | F1-Score |
|---|---|---|---|
| Normal | 0.92 | 0.91 | 0.92 |
| DDoS | 0.94 | 0.99 | 0.97 |
| DoS | 0.99 | 0.97 | 0.98 |
| Injection | 0.98 | 0.96 | 0.97 |
| Malware | 0.99 | 1.00 | 0.99 |
| MITM | 0.99 | 0.99 | 0.99 |
| Password | 0.99 | 0.97 | 0.98 |
| Probing | 0.99 | 0.99 | 0.99 |
| **Weighted Avg.** | **0.97** | **0.97** | **0.97** |

The ensemble model achieved an overall accuracy of 97% on the resampled test set, with detailed metrics shown in Table V. Notably, it maintained near-perfect F1-scores across most classes, demonstrating the benefit of aggregating complementary classifiers. This validates the choice of ensemble architecture as a robust base for later neutrosophic enhancements.

### C. Neutrosophic Scoring and Indeterminacy Distribution

To enhance explainability and quantify predictive uncertainty, we applied neutrosophic scoring to the ensemble output. Each prediction was decomposed into three components:

- T (Truth): Confidence in the predicted class
- F (Falsity): Aggregated confidence in all other classes
- I (Indeterminacy): Quantified ambiguity, derived from entropy across class probabilities

Using the averaged probability distributions from the ensemble, we computed neutrosophic scores for all test instances. The indeterminacy score $I$ is normalized entropy, scaled between 0 and 1. Values near 0 reflect confident predictions, while values near 1 indicate uncertainty or low consensus among classifiers.

The overall distribution of indeterminacy scores is shown in Figure 3. While most predictions exhibit low to moderate $I$-values, a significant portion exceed $I>0.4$, highlighting the presence of ambiguous samples within the test set.

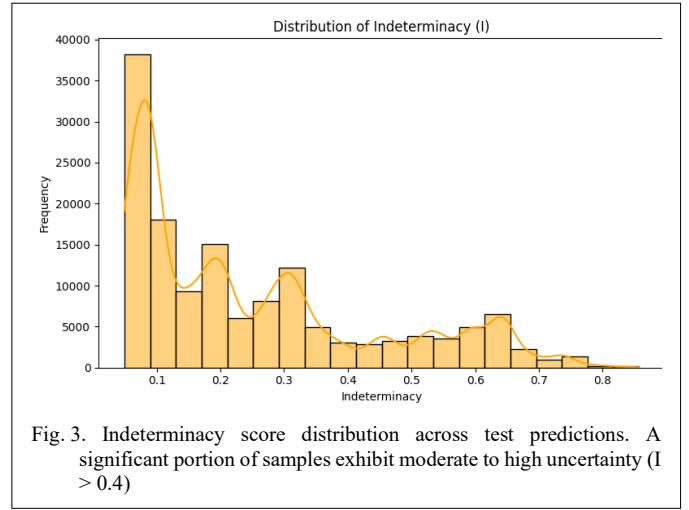

Fig. 3. Indeterminacy score distribution across test predictions. A significant portion of samples exhibit moderate to high uncertainty ($I > 0.4$)

To analyze uncertain cases, we flagged all predictions with $I > 0.4$. As shown in Figure 4, many high-indeterminacy samples clustered in the Normal, DDoS, and DoS classes, suggesting overlapping or subtle patterns that blur the line between benign and malicious behavior—especially in stealthy or obfuscated attacks.

In total, over 12,000 predictions were identified as high-indeterminacy. A substantial fraction of these were misclassified, affirming that $I$ serves as a robust proxy for epistemic uncertainty. This reinforces its value in driving adaptive abstention strategies and enabling more trustworthy intrusion detection in real-world, high-stakes environments.

### D. Threshold Sweep and Youden Index

To evaluate the trade-off between prediction confidence and decision coverage, we performed a sweep over indeterminacy thresholds $I \in [0.1, 0.9]$, progressively filtering out high-uncertainty predictions. For each threshold, we calculated:
- Accuracy on retained predictions ($I \leq \tau$)
- Coverage (fraction of samples retained),
- Youden Index (Accuracy × Coverage), which balances predictive performance and coverage.

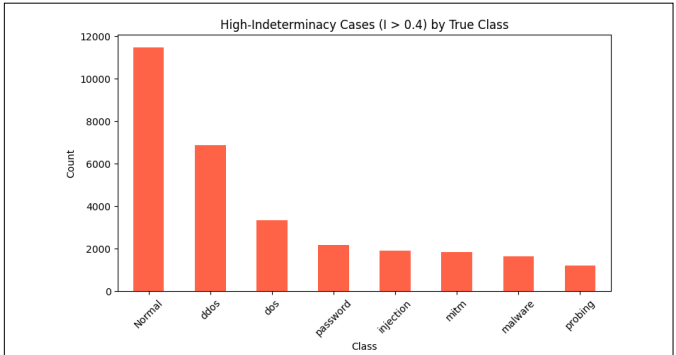

Fig. 4 Class distribution of high-indeterminacy predictions (I > 0.4), dominated by Normal, DDoS, and DoS

As shown in Figure 5, accuracy increases as the model abstains on higher-indeterminacy predictions, while coverage naturally decreases. The Youden Index peaked at $I \leq 0.8$, indicating the best balance between accuracy and decision coverage.

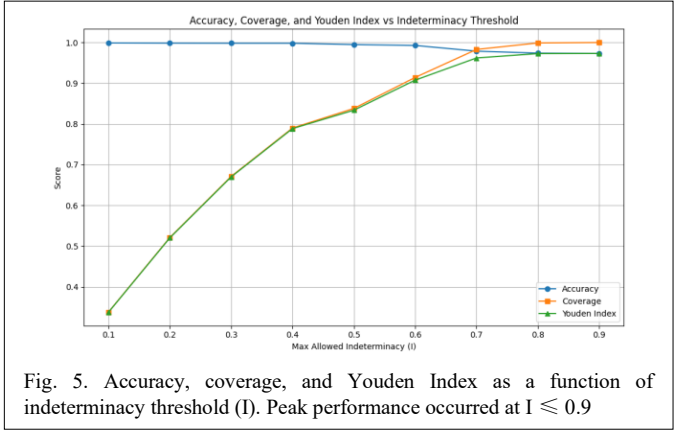

Fig. 5. Accuracy, coverage, and Youden Index as a function of indeterminacy threshold (I). Peak performance occurred at $I \leq 0.9$

At $I \leq 0.9$, the ensemble achieves 97.31% accuracy with full coverage, reflecting performance when no abstention is applied. However, abstaining on predictions with higher indeterminacy—such as filtering at $I \leq 0.4$—yields even higher precision, reaching over 99.8% accuracy on confident samples. This trade-off, visualized in Figure 5, shows how the system can prioritize interpretability and minimize risk in safety-critical scenarios by selectively deferring uncertain cases. Additionally, adaptive abstention based on per-class indeterminacy thresholds flagged approximately 20% of each class's samples for potential review, aligning with the 80th percentile thresholding strategy.

### E. Adaptive Abstention & Flagging

Beyond using a global indeterminacy threshold, NeutroSENSE supports class-specific adaptive abstention, which tailors uncertainty handling to the unique distributional properties of each class. This is motivated by the observation that certain categories (e.g., Normal, MITM) are inherently more prone to ambiguity due to overlapping behaviors or polymorphic signatures.

To implement this, we computed the 80th percentile indeterminacy score $I$ within each class on the test set. Predictions exceeding their class-specific threshold were flagged for potential human review. This approach enhances abstention sensitivity for hard-to-classify categories while minimizing unnecessary review for more separable classes.

As shown in Figure 6, the Normal, DDoS, and Malware classes exhibited broader and more variable indeterminacy scores, reflecting increased uncertainty in prediction. Meanwhile, Figure 4 highlights that Normal, DDoS, and DoS contributed the most high-indeterminacy samples ($I > 0.4$). Together, these insights suggest that traffic patterns in these classes—due to obfuscation or benign-like behavior—pose challenges for classification. NeutroSENSE's class-aware abstention helps prioritize human oversight for such ambiguous cases, improving trust and operational reliability in real-world deployments.

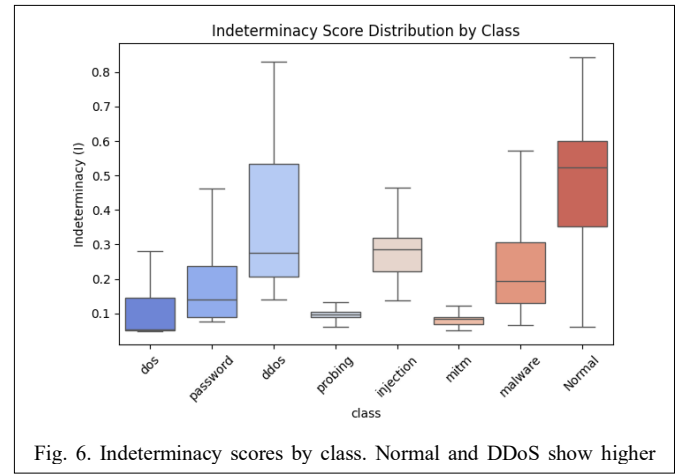

Fig. 6. Indeterminacy scores by class. Normal and DDoS show higher

To validate indeterminacy as a confidence signal, we analyzed its relationship to prediction correctness. As shown in Figure 7, misclassified samples had significantly higher indeterminacy scores, indicating greater model uncertainty. This supports our hypothesis that indeterminacy ($I$) is a strong proxy for epistemic uncertainty, helping identify likely errors. Table VI reinforces this, with an average $I$ of 0.6223 for incorrect and 0.2410 for correct predictions. This clear separation validates neutrosophic decomposition for uncertainty modeling, abstention, and trust-aware decision-making.

TABLE VI. AVERAGE INDETERMINACY BY PREDICTION CORRECTNESS.

| Prediction | Average Indeterminacy (I) |
|---|---|
| Incorrect | 0.6223 |
| Correct | 0.2410 |

### F. Discussion

We demonstrate that the neutrosophic-enhanced ensemble achieves strong predictive performance, with 97% overall accuracy (Table V) and nearly 99.8% accuracy on low-indeterminacy predictions (Figure 5). Misclassified samples exhibit significantly higher indeterminacy scores (Figure 7,

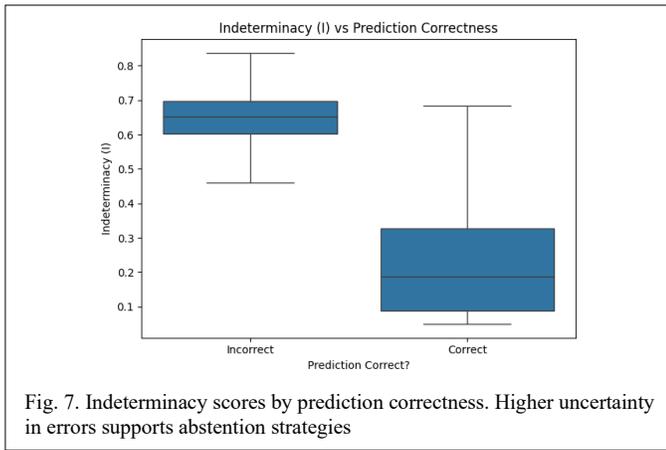

Fig. 7. Indeterminacy scores by prediction correctness. Higher uncertainty in errors supports abstention strategies

Table VI), confirming the effectiveness of neutrosophic decomposition for modeling uncertainty and enabling trust-aware abstention.

These results underscore the value of integrating neutrosophic logic into ensemble-based intrusion detection. By decomposing predictions into truth, falsity, and indeterminacy, NeutroSENSE can flag ambiguous cases and defer uncertain decisions—crucial for edge and IoT environments where automation risks are high. Abstention based on indeterminacy (Figure 5) improves interpretability, while adaptive thresholds (Figure 6) enable class-specific flagging. Although we did not benchmark against other uncertainty methods (e.g., entropy, margin), our findings show that neutrosophic reasoning enables behaviors beyond conventional ensembles—such as interpretable abstention and selective review of ambiguous cases—positioning NeutroSENSE as a practical foundation for trustworthy AI in cybersecurity.

## V. Conclusion

This paper introduced NeutroSENSE, a modular intrusion detection framework that integrates ensemble classification with neutrosophic reasoning to quantify and manage uncertainty. By combining prediction decomposition, indeterminacy scoring, and abstention, the framework enables more interpretable and trustworthy decision-making. Evaluation on an IoT dataset demonstrated high accuracy, effective detection of ambiguous cases, and support for class-specific uncertainty thresholds. Neutrosophic reasoning proved essential for capturing prediction uncertainty and improving operational reliability. Future work will focus on real-time deployment and adaptive threshold tuning in edge environments.